\documentclass[sigconf]{aamas}  
\usepackage{balance}  

\usepackage{booktabs}

\setcopyright{ifaamas}  
\acmDOI{}  
\acmISBN{}  
\acmConference[AAMAS'19]{Proc.\@ of the 18th International Conference on Autonomous Agents and Multiagent Systems (AAMAS 2019)}{May 13--17, 2019}{Montreal, Canada}{N.~Agmon, M.~E.~Taylor, E.~Elkind, M.~Veloso (eds.)}  
\acmYear{2019}  
\copyrightyear{2019}  
\acmPrice{}  

\usepackage{dsfont, amsfonts}
\usepackage{amsmath}
\DeclareMathOperator*{\argmax}{argmax}
\usepackage[colorinlistoftodos]{todonotes}
\usepackage{subfigure}
\usepackage{booktabs}
\usepackage{balance}

\begin{document}

\title{Using Causal Analysis to Learn Specifications\\ from Task Demonstrations}  


\author{Daniel Angelov}
\affiliation{%
 \institution{University of Edinburgh}
}
\email{d.angelov@ed.ac.uk}
\author{Yordan Hristov}
\affiliation{%
 \institution{University of Edinburgh}
}
\email{y.hristov@ed.ac.uk}

\author{Subramanian Ramamoorthy}
\affiliation{%
 \institution{University of Edinburgh}}
\email{s.ramamoorthy@ed.ac.uk}

\begin{abstract}  
Learning models of user behaviour is an important problem that is broadly applicable across many application domains requiring human-robot interaction. In this work we show that it is possible to learn a generative model for distinct user behavioral types, extracted from human demonstrations, by enforcing clustering of preferred task solutions within the latent space. We use this model to differentiate between user types and to find cases with overlapping solutions. Moreover, we can alter an initially guessed solution to satisfy the preferences that constitute a particular user type by backpropagating through the learned differentiable model. An advantage of structuring generative models in this way is that it allows us to extract causal relationships between symbols that might form part of the user's specification of the task, as manifested in the demonstrations. We show that the proposed method is capable of correctly distinguishing between three user types, who differ in degrees of cautiousness in their motion, while performing the task of moving objects with a kinesthetically driven robot in a tabletop environment. Our method successfully identifies the correct type, within the specified time, in 99\% $[97.8 - 99.8]$ of the cases, which outperforms an IRL baseline. We also show that our proposed method correctly changes a default trajectory to one satisfying a particular user specification even with unseen objects. The resulting trajectory is shown to be directly implementable on a PR2 humanoid robot completing the same task.

\end{abstract}

%


\maketitle


\section{Introduction}

\begin{figure}[t]
\centering
\includegraphics[width=1\linewidth]{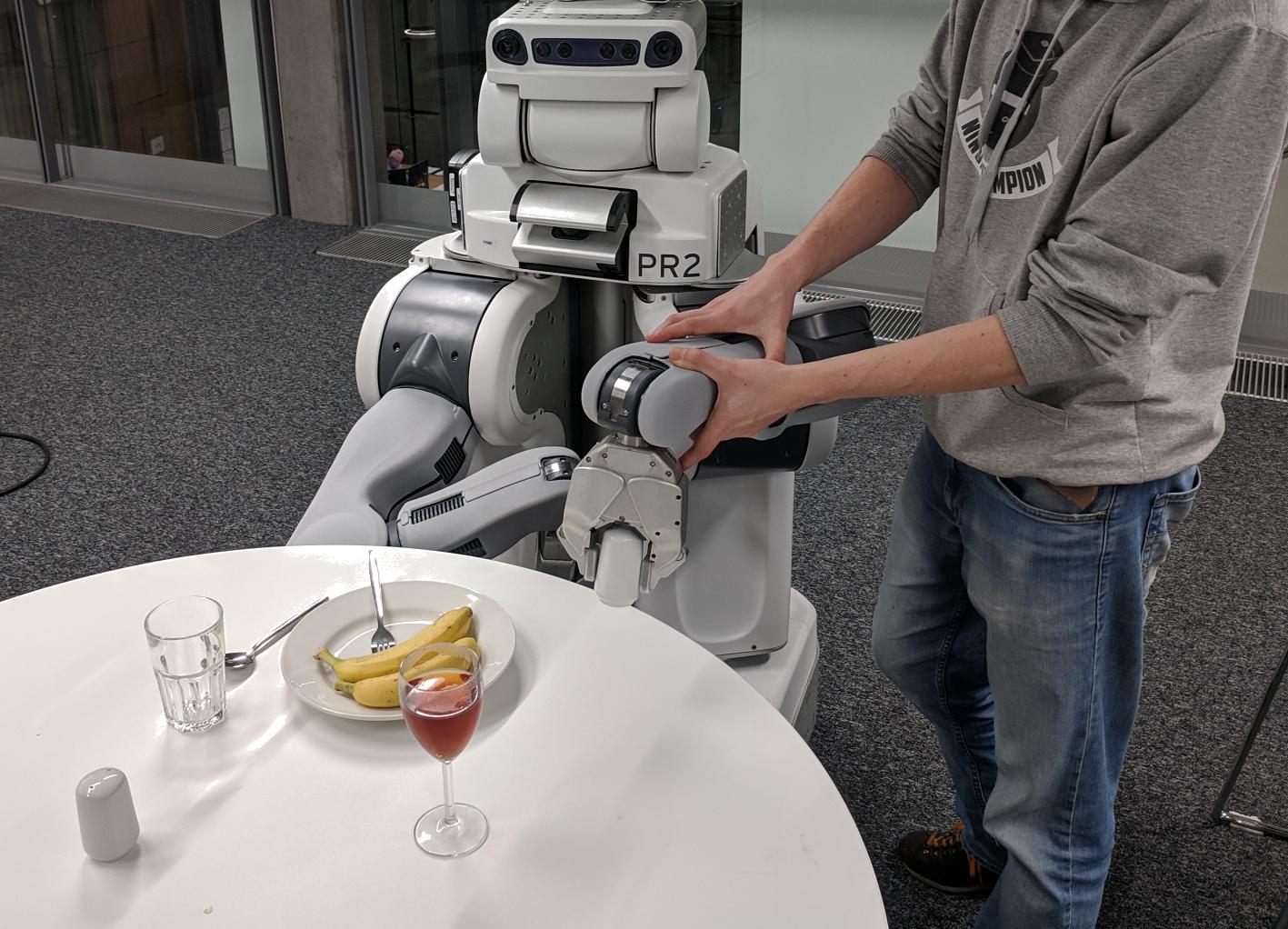}
\caption{Example setup - task is to return the pepper shaker to its original location---next to the salt shaker. Deciding which objects to avoid when performing the task can be seen as conditioning on the user specifications, implicitly given during a demonstration phase.}
\label{fig:robot_scene}
\end{figure}

As we move from robots dedicated to specific pre-programmed tasks to more general purpose tasks, there is a need for easy re-programmability of these robots. A promising approach to such easy re-programming is Learning from Demonstration, i.e., by enabling the robot to mimic behaviors shown to it by a human expert---Figure.~\ref{fig:robot_scene}. 

\begin{figure}[t]
\centering
\includegraphics[width=1.\linewidth]{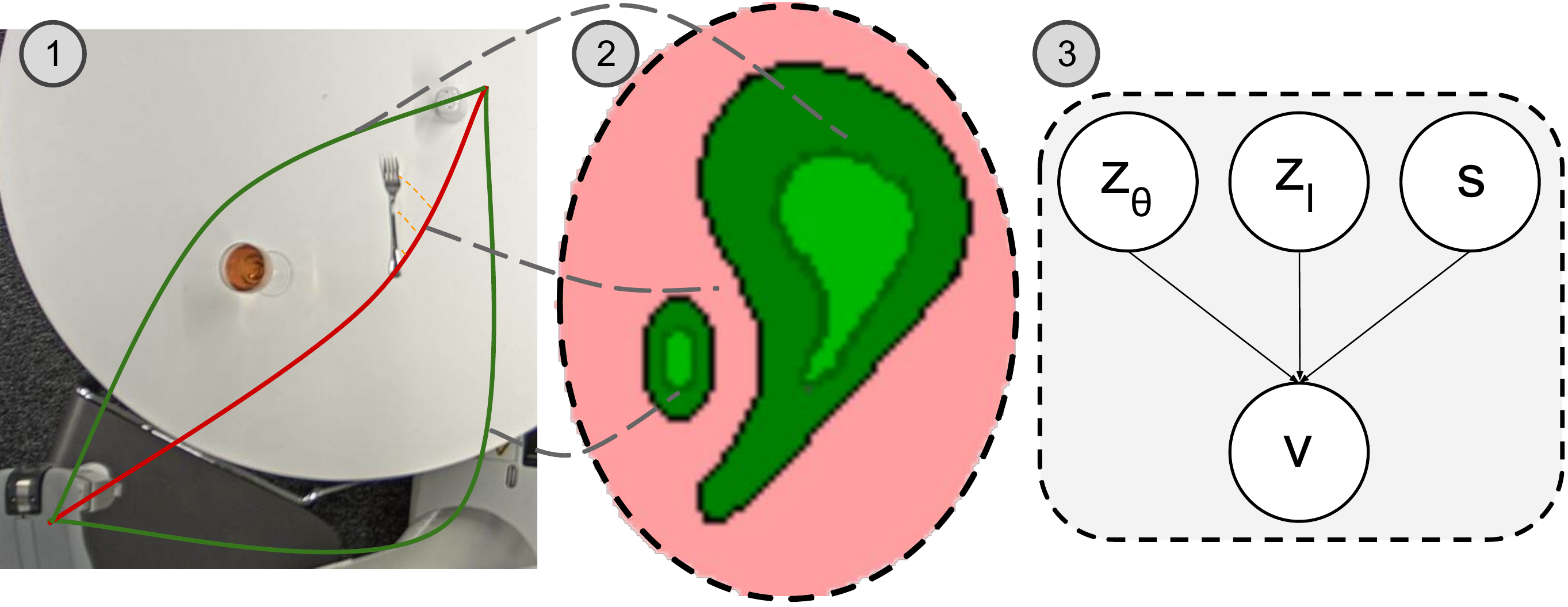}
\caption{ \textbf{(1)} Demonstrations that satisfy the user task specification maintain a distance from fragile objects (i.e. a wine glass), or fail to satisfy the specification by moving over sharp items. \textbf{(2)} An environment can have multiple clusters of valid trajectories in the latent space, conditioned on user type. \textbf{(3)} The validity of trajectories can be represented as a causal model. Whether a trajectory is part of a cluster $v$ is conditioned on the specific path $z_\theta$, the environment $z_I$, and the specification $s$.}
\label{fig:claims}
\end{figure}

With such a setup we can escape from having to handcraft rules and allow the robot to learn by itself, including modelling the specifications the teacher has used during the demonstration. Often some of these innate preferences are not explicitly articulated and are mostly biases resulting from experiences with other unrelated tasks sharing parallel environmental corpora~- Figure.~\ref{fig:claims}.1. The ability to notice, understand and reason causally about these deviations, whilst learning to perform the shown task is of significant interest.

Similarly, other methods for Learning from Demonstration as discussed by \citet{Argall2009} and \citet{wirth2017survey} in the Reinforcement Learning domain are focused on finding a general mapping from observed state to an action, thus modeling the system or attempting to capture the high-level user intentions into a plan. The resulting policies are not generally used as \textit{generative models}. As highlighted by \citet{Sunderhauf2018} one of the fundamental challenges with robotics is the ability to reason about the environment, beyond a state-action mapping.

Thus, when receiving a positive demonstration, we should aim to understand the causal reasons differentiating it from a non-preferential one, rather than merely mimicking the particular trajectory.
When people demonstrate a movement associated with a concept, they rarely mean one singleton trajectory alone. Instead, that instance is an element of a set of trajectories that share particular features. 
In other words, we want to find groups of trajectories with similar characteristics that may be represented as clusters in a suitable space. We are interested in learning these clusters so that subsequent new trajectories can be classified according to whether they are good representatives of the class of intended feasible behaviors.

It is often the case that in problems that exhibit great flexibility in  possible solutions, different experts may generate solutions that are part of different clusters - Figure.~\ref{fig:claims}.2. In cases where we naively attempt to perform statistical analysis, we may end up collapsing to a single mode or merging the modes in a manner that doesn't entirely reflect the underlying semantics (e.g., averaging trajectories for going left/right around an object).

We present a method for introspecting in the latent space of a model which allows us to relax some of the assumptions illustrated above and more concretely to:
\begin{itemize}
\item find varied solutions to a task by sampling a learned generative model, conditioned on a particular user specification.
\item backpropagate through the model to change an initially guessed solution towards an optimal one with respect to the user specification of the task.
\item counterfactually reason about the underlying feature preferences implicit in the demonstration, given key environmental features, and to build a causal model describing this.
\end{itemize}

When we talk about task specification, we understand the symbolic/
high-level descriptions of a desired behavior. The specifications, as learned by the network, are the observed regularities in the human behavior. 

Our proposed method relies on generating a latent space from environmental observations with the demonstrator's trajectories. This teacher's positive and negative examples are used as a guide for estimating the specification and validity of the trajectory parameterization.

\section{Related Work}
\subsection{Learning from Demonstration}
Learning from demonstration involves a variety of different methods for approximating the policy. In some related work, the state space is partitioned and the problem is viewed as one of classification. This allows for the environment state to be in direct control of the robot and to command its discrete actions - using Neural Networks (\citet{Matari1999}), Bayesian Networks (\citet{Inamura2000}), Gaussian Mixture Models (\citet{Chernova2007}). 
Alternatively, it can be used to classify the current step in a high-level plan \citet{Thomaz2004} and execute predetermined low-level control.

In cases where a continuous action space is preferred, regressing from the observation space can be achieved by methods like Locally Weighted Regression \citet{ClevelandSmoothingBL}. 

Roboticists e.g., \citet{Sunderhauf2018},  have long held the view that reasoning as part of planning is dependent on reasoning about objects, semantics and their geometric manifestations.
This is based on the view that structure within the demonstration should be exploited to better ground symbols between modalities and to the plan. 

One way to learn such latent structure can be in the form of a reward function through the usage of Inverse Reinforcement Learning methods as presented by \citet{ng2000algorithms, zhifei2012survey, brown2018machine}. However, it is not always clear that the underlying reward, which an expert may have used, is entirely re-constructable or even if it can be sufficiently approximated. Alternatively, preference-based reinforcement learning (PbRL), \citet{wirth2017survey}, offers methods whose focus lies in learning from non-numeric rewards, directly from the guidance of a demonstrator. Such methods are particularly useful for problems with high-dimensional domains, e.g. robotics - \citet{jain2013learning, jain2015learning}, where a numeric reward (unless highly shaped) might not be able to capture all semantic subtleties and variations contained in the expert's demonstration. Thus, in the context of PbRL, the method we propose learns a user specification model using user-guided exploration and trajectory preferences as a feedback mechanism, using definitions from \citet{wirth2017survey}. 

\subsection{Causality and State Representation}
The variability of environmental factors makes it hard to build systems relying only on correlation data statistics for specifying their state space. Methods that rely on causality, \citet{Pearl2009, Harradon2018}, and learning the cause and effect structure, \citet{Rojas2017}, are much better suited to support the reasoning capabilities for transferring core knowledge between situations. Interacting with the environment allows robots to perform manipulations that can convey new information to update the observational distribution or change their surrounding and in effect perform interventions within the world. 
Performing counterfactual analysis helps in a multi-agent situation with assignment of credit as shown by \citet{foerster2017counterfactual}.

Learning sufficient state features has been highlighted by \citet{Argall2009} as a future challenge for the LfD community. The problem of learning disentangled representations aims to generate a good composition of a latent space, separating the different modes of variation within the data. 
\citet{higgins2016beta, Chen2018} have shown promising improvements in disentangling of the latent space with minimal or no assumption by manipulating the Kullback - Leibler divergence loss of a variational auto-encoder. 
\citet{Denton2017} show how the modes of variation for content and temporal structure should be separated and can be extracted to improve the quality of the next frame video prediction task if the temporal information is added as a learning constraint.  
While the disentangled representations may not directly correspond to the factors defining action choices, \citet{Johnson2016} adds a factor graph and composes latent graphical models with neural network observation likelihoods. 

The ability to manipulate the latent space and separate variability as well as obtain explanation about behavior is also of interest to the interpretable machine learning field, as highlighted by \citet{Doshi2017}.


\section{Problem Formulation}
\label{sec:problem_formulation}
In this work, we assume that the human expert and robotic agent share multiple static tabletop environments where both the expert and the agent can fully observe the world and can interact with an object being manipulated. The agent can extract RGB images of static scenes and can also be kinesthetically driven while a demonstration is performed. The task at hand is to move an object held by the agent from an initial position $p_{init}$ to a final position $p_f$ on the table, while abiding by certain user-specific constraints. Both $p_{init}$ and $p_f \in \mathbb{R}^P$. The user constraints are determined by the demonstrator's type $s$, where $s \in S=\{s_1, \ldots, s_n\}$ for $n$ user types.

Let $\mathcal{D} = \{\{\mathbf{x}_1, v_1\}, \ldots, \{\mathbf{x}_N, v_N\}\}$ be a set of N expert demonstrations, where $\mathbf{x}_i = \{I, tr_{i}^{s}\}$, $I \in \mathbb{R}^M$ is the tabletop scene, $tr_{i}^{s}$ is the trajectory and $v_i$ is a binary label denoting the validity of the trajectory with respect to the user type $s$. 
Each trajectory $tr_{i}^{s}$ is a sequence of points $\{p_0, \ldots, p_{T_{i}}\}$, where $p_0 = p_{init}$ and $p_{T_{i}} = p_f$. The length of the sequences is not constrained---i.e. $T$ is not necessarily the same for different trajectories. 

The learning task is to project each $\mathbf{I} \in \mathbb{R}^M$ into $\mathbf{Z_I} \in \mathbb{R}^{K}$, $Z_I = E(I)$, and $tr_{i}^{s} \in \mathbb{R}^{PT_{i}}$ into $\mathbf{Z_\theta} \in \mathbb{R}^L$, $Z_\theta = Bz(tr^s_i)$, $K \ll M$, $L \ll PT_{i}$. Both $Z_I$ and $Z_\theta$ are used in order to predict the validity $\hat{v}_i, \hat{v}_i = C_s(Z_I, Z_\theta)$ of the trajectory $tr_{i}^{s}$ with respect to the user type $s$. With optimally-performing agent, $\hat{v}_i \equiv v_i$. For more details see Figure \ref{fig:arch}.

In order to alter an initial trajectory, we can find the partial derivative of model with respect to the trajectory parameters with the model conditioned on a specific user type $s$,

\[\Delta = \frac{\partial{C_s(z|\hat{v}=1)}}{\partial{z_{\theta}}} \]

We can take a gradient step $\Delta$ and re-evaluate. Upon satisfactory outcome, we can reproject back $z_\theta$ to a robotic executable trajectory $tr^s = Bz^{-1}(z_\theta)$.

The main capability we want from our model is to structure the latent space in a way that would not only allow us to distinguish between trajectories conforming and not to the user specifications and in turn generate good trajectories, but also to maintain variability in order to estimate the causal link between the symbols within the world and the validity of a trajectory, given some specification.

\section{Specification Model}

\begin{figure}[h]
\centering
\includegraphics[width=1\linewidth]{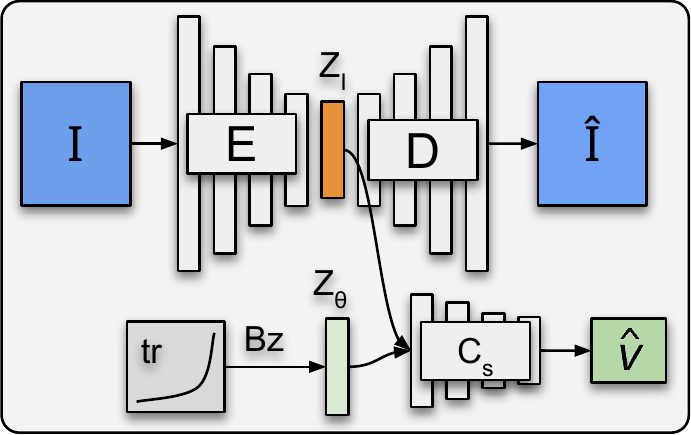}
\caption{Specification model architecture. The environmental image \textit{I}, $I \in R^{100\times100\times3}$, is passed through an Encoder-Decoder Convolutional Network, with a $16-8-4$ 3x3 convolutions, followed by fully connected layer, to create a compressed representation \textit{$Z_I, Z_I \in R^{15}$}. It is passed along with the trajectory parameterization $Z_\theta, Z_\theta \in R^{2}$ through a 3-layer fully connected classifier network that checks the validity of the trajectory \textit{$C_s(z)$} with respect to the spec. $s$.}
\label{fig:arch}
\end{figure}

We use the Deep Variational Auto-Encoder Framework---see \citet{Kingma2013}---as a base architecture. The full model consists of a convolutional encoder network $q_\phi$, parametrised by $\phi$, a deconvolutional decoder network $p_\psi$, parametrised by $\psi$, and a classifier network $C$, comprised of a set of fully-connected layers. The encoder network is used to compress the world representation \textit{I} to a latent space \textit{$Z_I$}, disjoint from the parameterization of the trajectories \textit{$Z_\theta$}. The full latent space is modeled as the concatenation of the world space and trajectory space \textit{$Z = Z_I \cup Z_\theta$} as seen on Figure.~\ref{fig:arch}.

Parameters---$\alpha, \beta, \gamma$---are added to the three terms of the overall loss function---see Equation \ref{eq:lossf}---so that their importance during learning can be leveraged. In order to better shape the latent space and to coerce the encoder to be more efficient, the Kullback-Leibler divergence loss term is scaled by a $\beta$ parameter, as in \citet{higgins2016beta}.
\begin{align} \label{eq:lossf}
\min_{\psi, \phi, \mathbf{C}} \mathcal{L}(\psi, \phi; I, & z_I, z_\theta, v) = \\
 & - \alpha \mathds{E}_{q_{\phi} (z_I|I)}(log p_{\psi}(I|z_I)) \nonumber \\
 & + \beta D_{KL}(q_{\phi}(z_I|I) || p_{\psi}(z_I)) \nonumber \\
 & - \gamma \left[v \log(C(z)) + (1-v) \log(1-C(z)) \right] \nonumber
\end{align}

By tuning its value we can ensure that the distribution of the latent projections in $Z_I$ do not diverge from a prior isotropic normal distribution and thus influence the amount of disentanglement achieved in the latent space. A fully disentangled latent space has factorised latent dimensions---i.e. each latent dimension encodes a single data-generative factor of variation. It is assumed that the factors are independent of each other. For example, one dimension would be responsible for encoding the X position of an objectin the scene, another for the Y position, third for the color, etc. \citet{higgins2018scan} and \citet{chen2016infogan} argue that such low-dimensional disentangled representations, learned from high-dimensional sensory input, can be a better foundation for performing separate tasks~-~trajectory classification in our case.
Moreover, we additionally add a binary cross-entropy loss (scaled by \textit{$\gamma$}) associated with the ability of the full latent space $Z$ to predict whether a trajectory $tr^s$ associated a world $I$ satisfies the semantics of the user type $s$ - $\hat{v}$. We hypothesise that by backpropagating the classification error signal through $Z_I$ would additionally enforce the encoder network to not only learn factorised latent representations that ease reconstruction, but also trajectory classification. The full loss can be seen in Eq.~\ref{eq:lossf}.

The values for the three coefficients were empirically chosen in a manner such that none of the separate loss terms overwhelms the gradient updates while optimising $\mathcal{L}$.

\section{Causal Modeling}

Naturally, our causal understanding of the environment can only be examined through the limited set of symbols, \textit{$O$}, that we can comprehend about the world. In this part, we work under the assumption that an object detector is available for these objects (as the focus of this work is on elucidating the effect of these objects on the trajectories rather than on the lower level computer vision task of object detection per se). Given this, we can construct specific world configurations to test a causal model and use the above-learned specification model as a surrogate to inspect the validity of proposed trajectories.

\begin{figure}[h]
\centering
\includegraphics[width=0.4\linewidth]{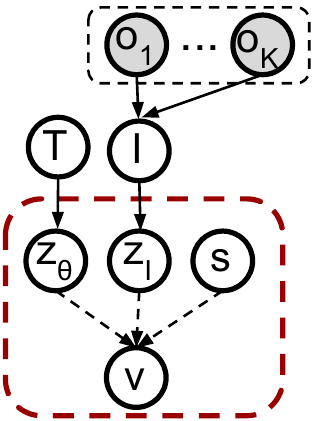}
\caption{The environment, compressed to \textit{$z_I$}, is composed of objects ($o_1, .., o_K$). A trajectory $T$ is parameterized by $z_\theta$, which alongside the factors \textit{$z_I$} and user specification \textit{s} are part of the specification model.}
\label{fig:scm}
\end{figure}

If we perform a search in the latent space \textit{$z_\theta$}, we can find boundaries of trajectory validity. We can intervene and counterfactually alter parameters of the environment and specifications and see the changes of the trajectory boundaries. By looking at the difference of boundaries in cases where we can test for associational reasoning, we can causally infer whether 

\begin{itemize}
    \item the different specifications show alternate valid trajectories
    \item a particular user specification reacts to the existence of a specific symbol within the world.
\end{itemize}.

\subsection{Specification Model Differences}

We are interested in establishing the causal relationship within the specification model as shown on Figure.~\ref{fig:scm}. We define our Structural Causal Model (SCM), following the notation of \citet{Peters2017} as

\[\mathfrak{C} := (\textbf{S}, P_\textbf{N}), ~S = \{\textbf{$X_j$} := \textit{f}_j(\textbf{PA}_j, N_j) \} \]

where nodes $\textbf{X} = \{Z_\theta, Z_I, S, V\}$ and $\textbf{PA}_j = \{ X_1, X_2, .. X_n \} \backslash \{X_j\} $. Given some observation \textbf{x}, we can define a counterfactual SCM
\( \mathfrak{C}_{\textbf{X}=\textbf{x}} := (\textbf{S}, P_\textbf{N}^{\mathfrak{C} | \textbf{X} = \textbf{x}}) \)
, where \( P_\textbf{N}^{\mathfrak{C} | \textbf{X} = \textbf{x}} := P_{N|\textbf{X}=\textbf{x}}\)

We can choose a particular user specification $s \sim p(S), s \neq s_\textbf{x} $ and use the specification model to confirm that the different specification models behave differently given a set of trajectories and scenes, i.e. the causal link $s \rightarrow v$ exists by showing: 

\begin{align}
\label{eq:neq1}\mathbb{E}\left[ P_v^{\displaystyle \mathfrak{C} | \textbf{X} = \textbf{x}} \right] \neq \mathbb{E}\left[ P_v^{\displaystyle \mathfrak{C}|\textbf{X}=\textbf{x}; do (S:=s)} \right] 
\end{align}

We expect different user types to generate a different number of valid trajectories for a given scene. Thus, by intervening on the user type specification we anticipate the distribution of valid trajectories to be altered, signifying a causal link between the validity of a trajectory within a scene to a specification.

\subsection{Symbol Influence on Specification Models}

We want to measure the response of the specification models of intervening in the scene and placing additional symbols within the world. We use the symbol types $O = \{o_1, .., o_k\}$ as described in Section.~\ref{sec:dataset}.
To accomplish this, for each symbol within the set we augment the scene $I$, part of the observation \textbf{x} with symbol $o$, such that $I_{new} = I \cup o$. If we observe that the entailed distributions of $P_v^{\mathfrak{C}|\textbf{X}=\textbf{x}; do (Z_I:=z_{Inew})}$ changes i.e.

\begin{align}
\label{eq:neq_symbol}\mathbb{E}\left[ P_v^{\displaystyle \mathfrak{C} | \textbf{X} = \textbf{x}} \right] \neq \mathbb{E}\left[ P_v^{\displaystyle \mathfrak{C}|\textbf{X}=\textbf{x}; do (Z_I:=z_{Inew})} \right] 
\end{align}

then the introduced object $o$ has a causal effect upon the validity of trajectories conditioned upon the task specification $s_\textbf{x}$.

We investigate the intervention of all symbol types permutated with all task specifications.

\section{Experimental Setup}

\subsection{Dataset}
\label{sec:dataset}

\begin{figure}[h]
\centering
\includegraphics[width=1\linewidth]{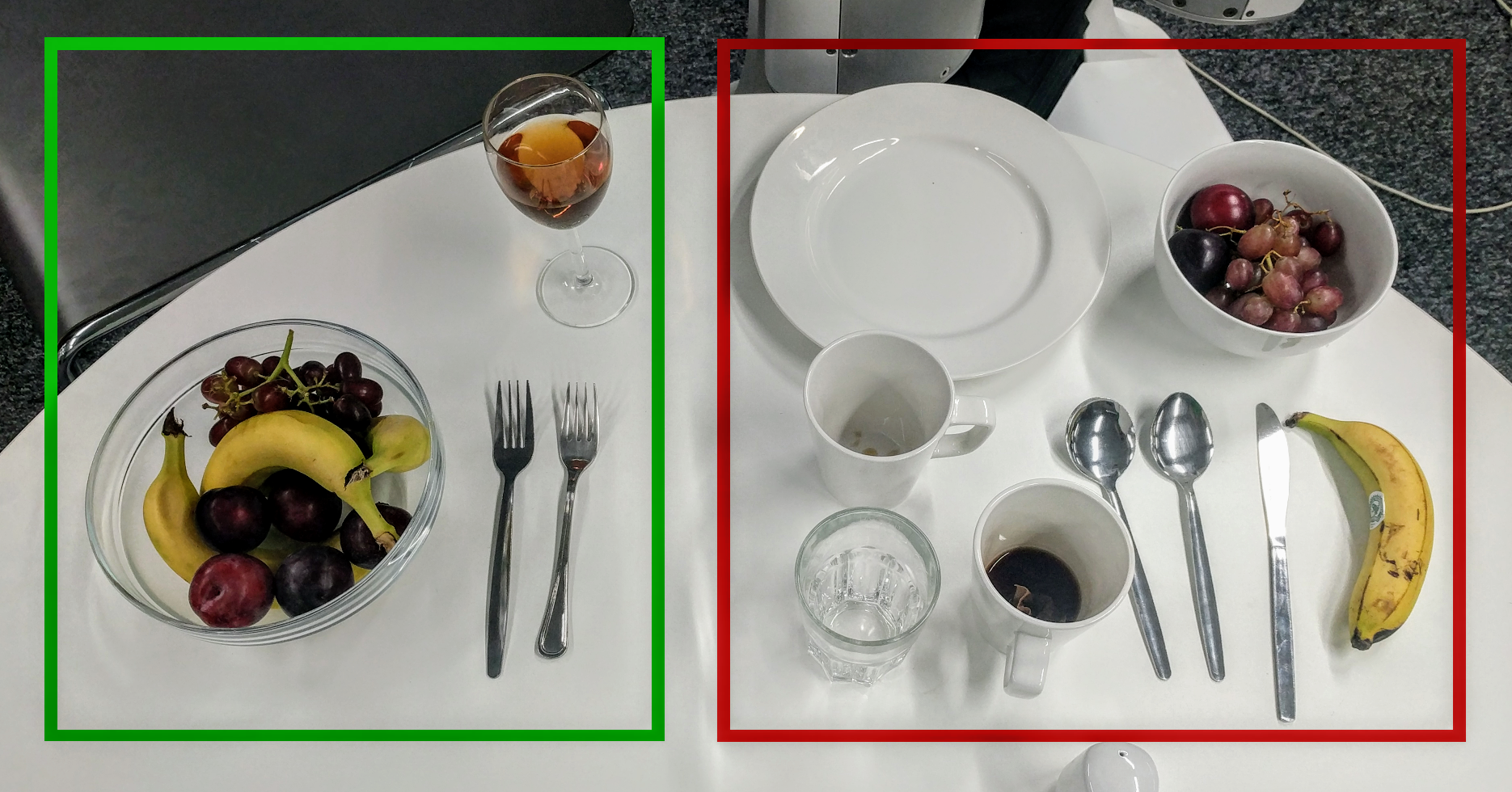}
\caption{Items used for the generation of the training (green) and test (red) scenes.}
\label{fig:train_test_items}
\end{figure}

\begin{figure}[h]
\centering
\includegraphics[width=1\linewidth]{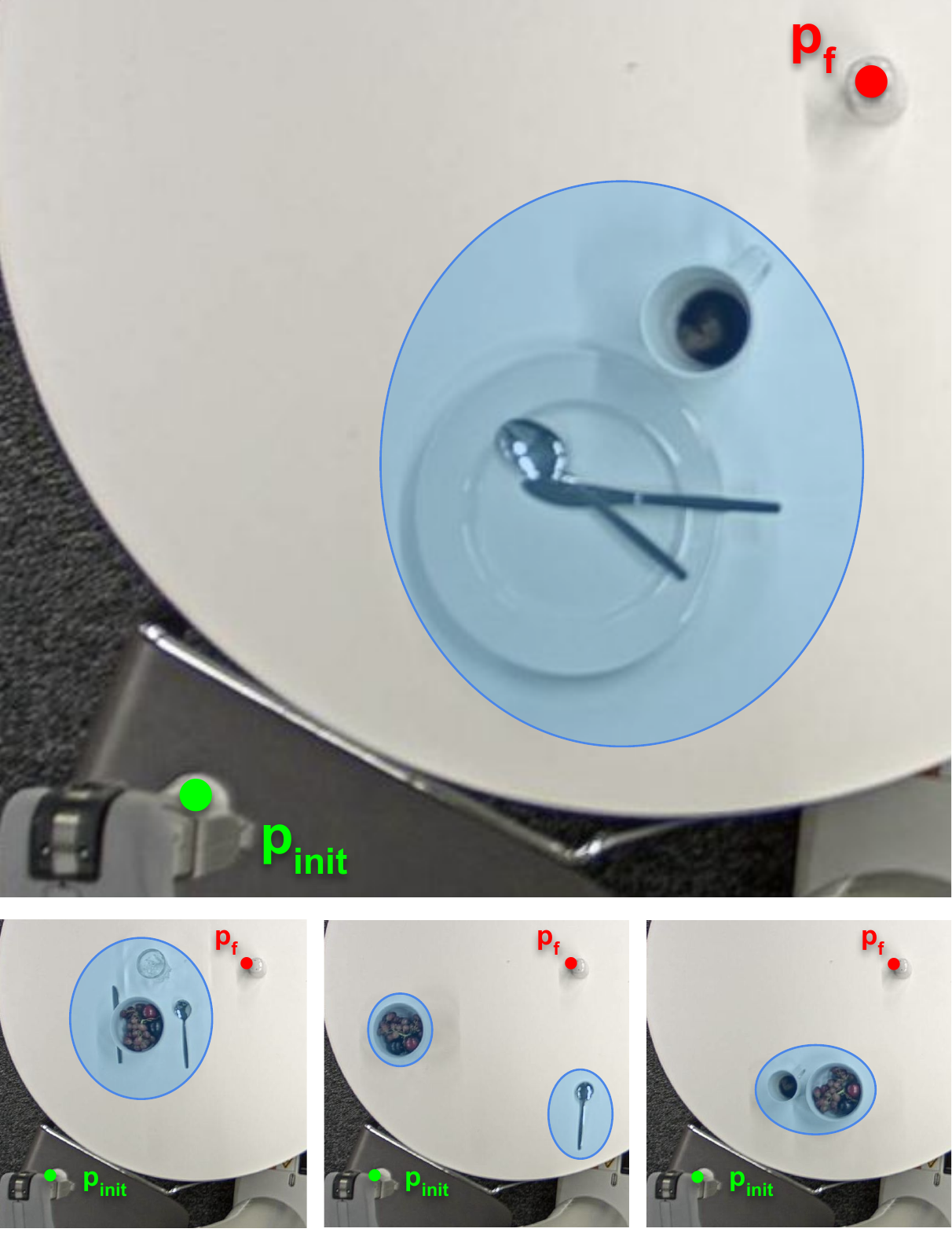}
\caption{Sample images used to represent example scenes. $p_{init}$ and $p_f$ are as defined in Section \ref{sec:problem_formulation}. Blue blobs represent potential obstacles in the scene, which some user types might want to avoid, and are only drawn for illustrative purposes.}
\label{fig:sample}
\end{figure}

The environment chosen for the experiment consists of a top down view of a tabletop on which a collection of items,~\textit{O=\{utensils, plates, bows, glasses\}} - Figure.~\ref{fig:train_test_items}, usually found in a kitchen environment, have been randomly distributed. 
The task that the demonstrator has to accomplish is to kinestetically move a robotic arm gently holding a pepper shaker from one end on the table to the other ($p_{init}=$bottom left, $p_f$=top right) by demonstrating a trajectory, whilst following their human preferences around the set of objects ---see Figure \ref{fig:sample}. The demonstrators are split into user types $S$, $ S=\{careful, ~normal, ~aggressive\}$ based on the trajectory interaction with the environment. The semantics behind the types are as follows: the \textit{careful} user tries to avoid going near any objects while carrying the pepper shaker, the \textit{normal} user tries to avoid only cups and the \textit{aggressive} user avoids nothing and tries to finish the task by taking the shortest path from $p_{init}$ to $p_f$. 

The agent observes the tabletop world and the user demonstrations in the form of 100x100 pixel RGB images $I, I \in \mathbb{R}^{100\times100\times3}$. The demonstrator---see Figure.~\ref{fig:robot_scene}---is assigned one of the types in $S$, has to produce a number of possible trajectories, some that satisfy the semantics of their type and some that break it - Figure.~\ref{fig:claims}.1. As specified in Section \ref{sec:problem_formulation}, each trajectory $tr^s$ is a sequence if points $\{p_0, \ldots, p_{T}\}$, where $p_0 = p_{init}$ and $p_{T_{i}} = p_f$. Each point $p_j, j \in \{0, \ldots, T\}$ represents the 3D position of the agent's end effector with respect to a predefined origin. However, all kinesthetic demonstrations are performed in a 2D (XY) plane above the table, meaning that the third coordinate of each point $p_j$ carries no information ($P=2$). An efficient way to describe the trajectories is by using a Bezier curve representation--- see \citet{bezier_curve}. The parameterization of a single trajectory becomes the 2D location of the central control point parametrized by $\theta$, together with $p_{init}$ and $p_f$. However, the initial and final points for each trajectory are the same and we can omit them. Thus, with respect to the formulations in Section \ref{sec:problem_formulation} $L=2$ and $Z_\theta \in \mathbb{R}^2$.

In total, for each user type $s \in S$, 20 scenes are used for training, with 10 trajectories per scene. The relationship between the number of trajectories per scene and the model's performance is explored in Section \ref{sec:results}. For evaluation purposes additional 20 scenes are generated, using a set of new items that have not been seen before---see Figure.~\ref{fig:train_test_items}.

\subsection{Evaluation}

We evaluate the performance of the model by its ability to correctly predict the validity of a trajectory with a particular specification. We perform an ablation study with the full model ($\alpha \neq 0, \beta \neq 0, \gamma \neq 0,$), AE model ($\beta = 0$), and classifier ($\alpha = 0, \beta = 0$). We investigate how the performance of the model over unseen trajectories varies with a different number of trajectories used for training per scene. We randomize the data used for training 10 times and report the mean.   

As a baseline we use an IRL model $r_s(p, I)$, such that the policy $\pi$ producing a trajectory $tr^s$ that is optimal wrt: 

\[\argmax_{tr^s} \sum_{i=0}^N{r_s(p_i, I)}\]

Additionally, we test the ability of the learned model to alter an initially suggested trajectory to a valid representative of the user specification. We assess this on the test set with completely novel objects by taking 30 gradient steps and marking the validity of the resulting trajectory.

We perform a causal analysis of the model with respect to the different user specifications and evaluate the difference in their expected behavior. Additionally, we intervene by augmenting the images to include specific symbols and evaluate the difference of the expectation of their entailed distribution. This highlights how the different specifications react differently to certain symbols.

\begin{figure}[ht]
\centering
\includegraphics[width=1.\linewidth]{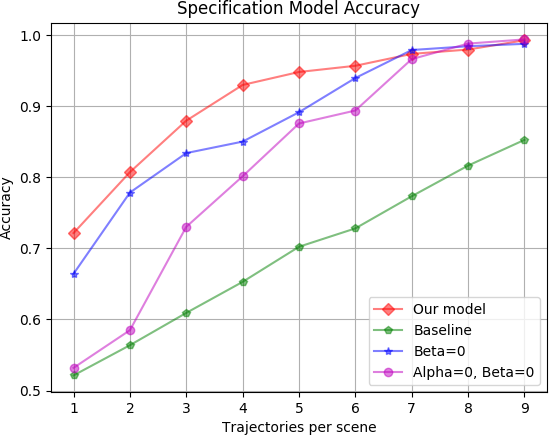}
\caption{The accuracy of the different models with respect the number of trajectories used within a scene. The lines indicate the mean accuracy with 10 different seed randomizations of the data. As the number of trajectories per scene increases, the performance of all models improves, but especially with a lower number of trajectories, our full model shows the biggest gains. }
\label{fig:model_perf}
\end{figure}

\begin{figure*}[ht]
\centering
\subfigure[Careful]{\includegraphics[width=0.32\linewidth]{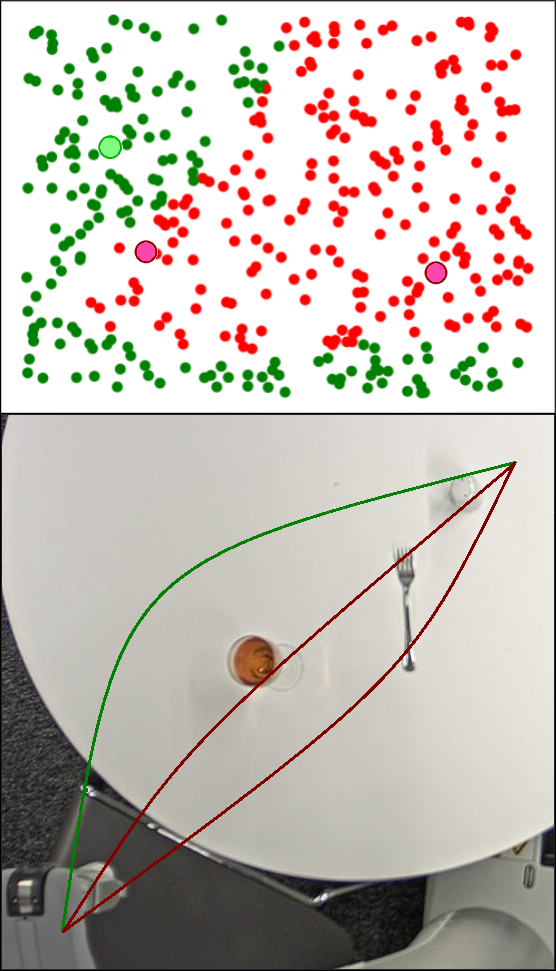}}
\hfill
\subfigure[Normal]{\includegraphics[width=0.32\linewidth]{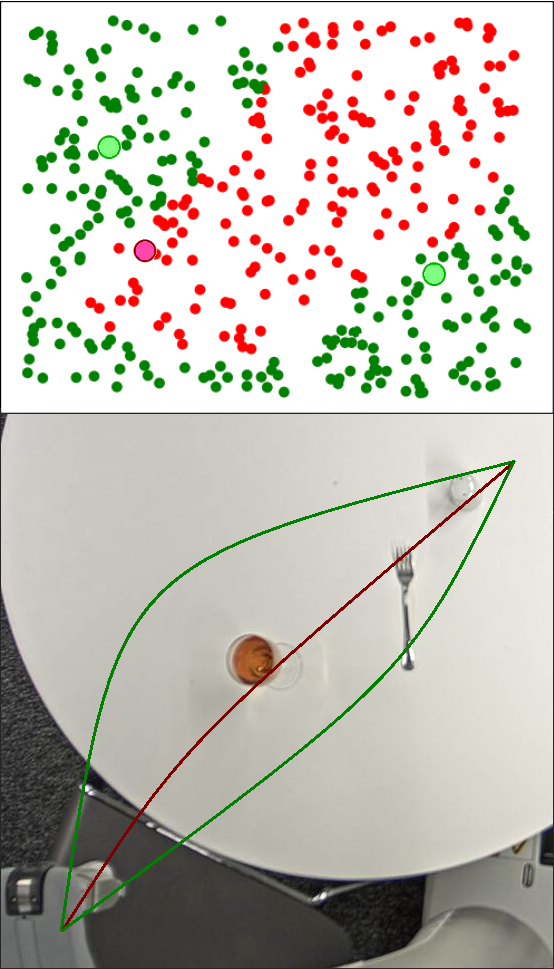}}
\hfill
\subfigure[Aggressive]{\includegraphics[width=0.32\linewidth]{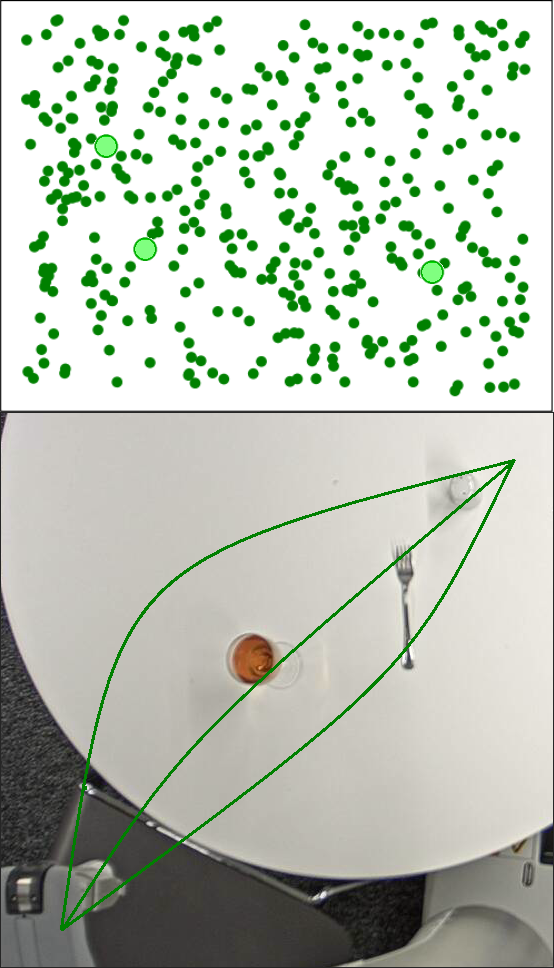}}
\caption{Sampling of the latent trajectory space---$Z_\theta$---of the preference model with different specifications. It can be observed how for the same region in the latent trajectory space---e.g. bottom right---the different user types have different validity values for the same trajectory---e.g. normal vs. careful user types around the cutlery and glass.}
\label{fig:types_traj}
\end{figure*}

\section{Results}
\label{sec:results}
In this section we illustrate how modeling the specifications of a human demonstrator's trajectories, in a table-top manipulation scenario within a neural network model, can be later used to infer causal links through a set of known features about the environment.

\subsection{Model Accuracy}
We show the accuracy the specification model on Figure.~\ref{fig:model_perf} and on our \href{https://sites.google.com/view/learnspecifications}{website}\footnote{Website on \url{https://sites.google.com/view/learnspecifications}}. Changing the number of trajectories shown within a scene has the highest influence on the performance ranging from $72\% [67.3 - 77.5]$ for a single trajectory to $99\% [97.8 - 99.8]$ \footnote{The numbers in brackets indicate the first and third quartile.} when using 9. The results illustrate that the models benefit from having an auto-encoder component to represent the latent space. However, they asymptotically approach perfect behavior as the number of trajectories per scene increases. Interestingly, the IRL baseline shows it needs much more information in order to create an appropriate policy.

If we look into the latent space of the trajectory --- Figure.~\ref{fig:types_traj} --- we can see that the trajectory preferences have clustered and there exists an overlap between the different model specifications. It also illustrates what the models' specifications can show about the validity of the trajectory.


\subsection{Trajectory Backpropagation}

We can use the learned specification model and perturb an initially suggested trajectory to suit the different user types by backpropagating through it and taking gradient steps within the trajectory latent space. 

\begin{figure*}[hbt]
\centering
\subfigure{\includegraphics[width=0.33\linewidth]{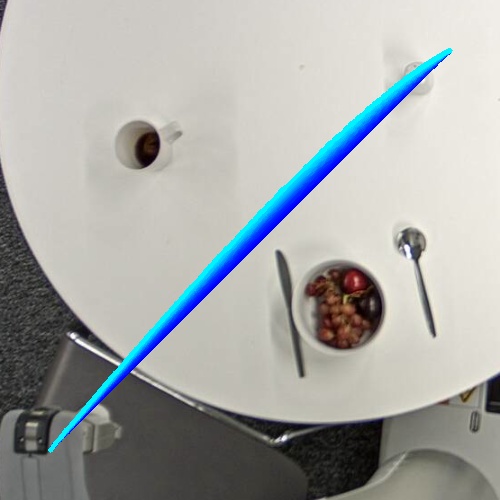}}
\subfigure{\includegraphics[width=0.33\linewidth]{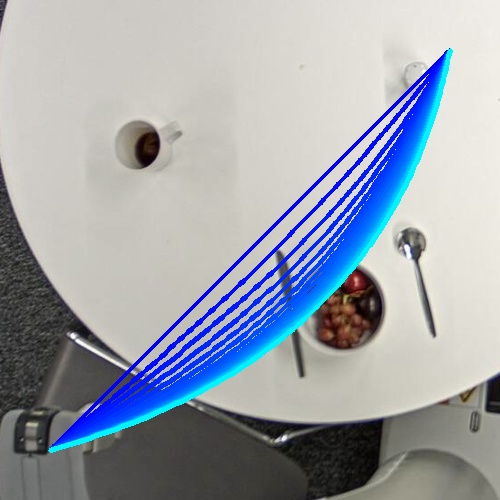}}
\subfigure{\includegraphics[width=0.33\linewidth]{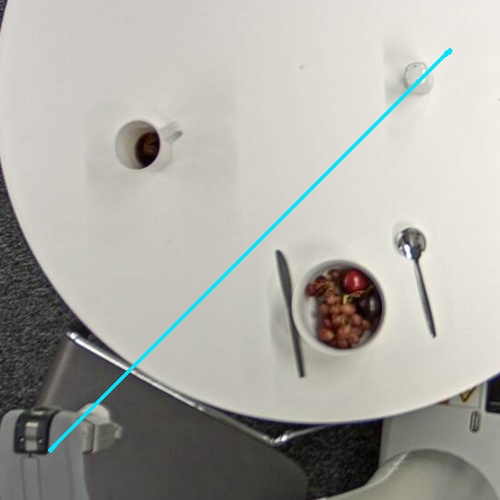}}

\vspace{-1em}

\subfigure{\includegraphics[width=0.33\linewidth]{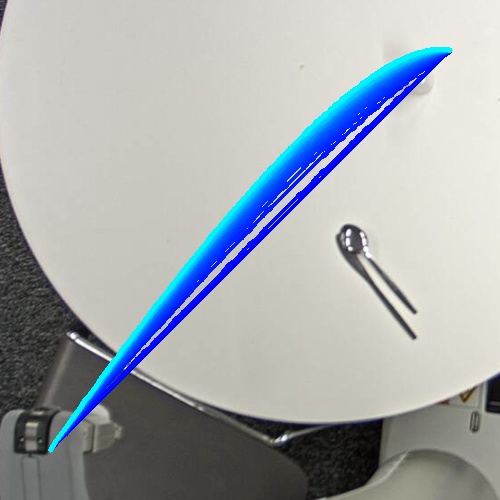}}
\subfigure{\includegraphics[width=0.33\linewidth]{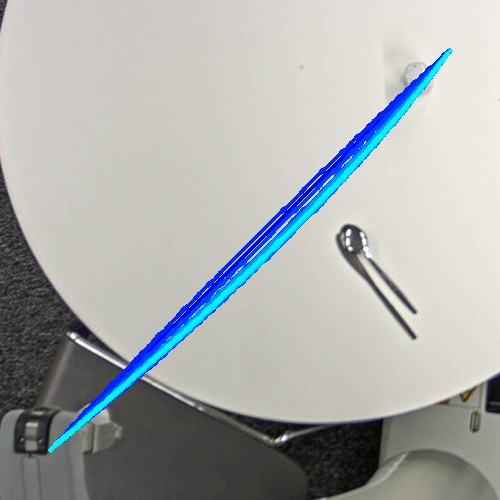}}
\subfigure{\includegraphics[width=0.33\linewidth]{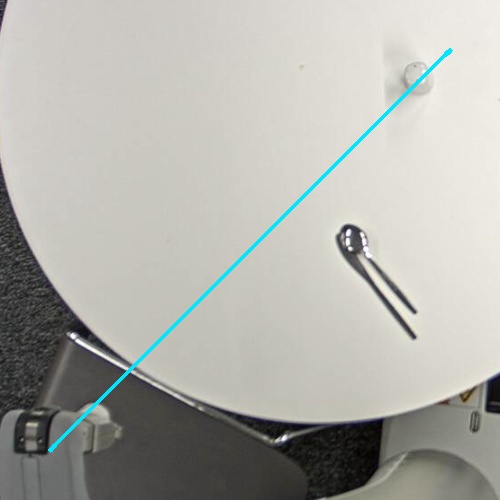}}

\vspace{-1em}

\setcounter{subfigure}{0}
\subfigure[Careful]{\includegraphics[width=0.33\linewidth]{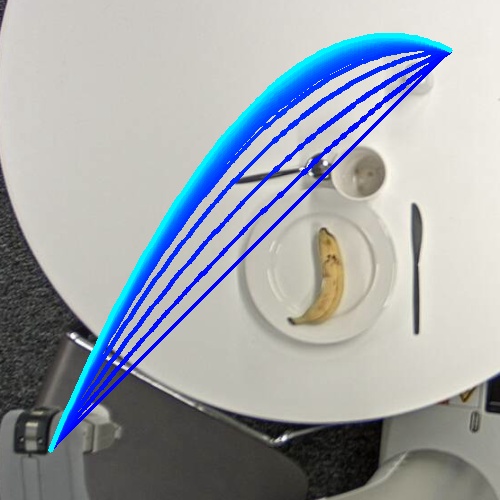}}
\subfigure[Normal]{\includegraphics[width=0.33\linewidth]{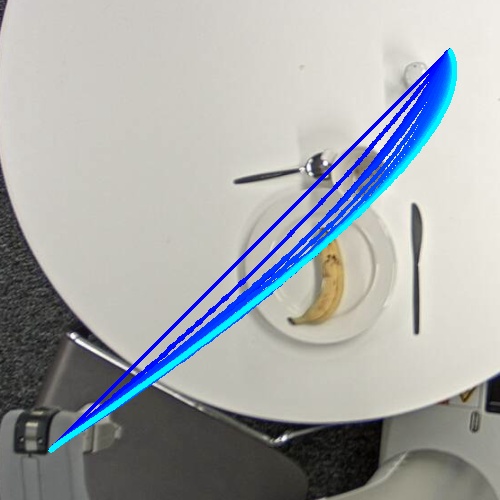}}
\subfigure[Aggressive]{\includegraphics[width=0.33\linewidth]{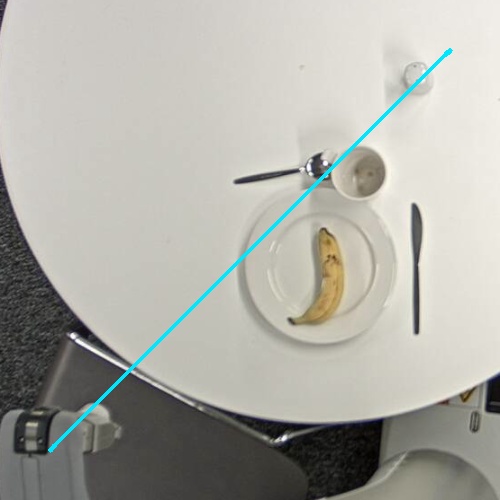}}

\caption{An initial trajectory (seen in dark blue) is used as a base solution to the task for difference scenes---rows 1, 2, 3. Furthermore, the parametrisation $z_\theta$ for each initial trajectory is continuously updated so that it better abides by the semantics of the different user specifications--- columns a,b,c. It can be seen that as the gradient steps in $Z_\theta$ are taken, the resulting intermediate trajectories are shifted to accommodate the preference of the model until the final trajectory (light blue) is reached. Color change from dark to light blue designates progressive gradient steps. \vspace{0.5em}}
\label{fig:packprop}
\end{figure*}

Based on the unseen object test scenes, the models were evaluated under the different specifications and the results can be found in Table.~\ref{table:backprop}. Individual trajectory movements can be seen in Figure.~\ref{fig:packprop}.

The first row of Figure.~\ref{fig:packprop} shows that the careful user type steering away from both the cup and bowl/cutlery, whereas in the normal user type, the model prefers to stay as far away from the cup as possible, ignoring the bowl. The model conditioned on the aggressive user type does not alter its preference of the trajectory, regardless of it passing through objects. The second row illustrates a situation, where the careful model shifts the trajectory to give more room to the cutlery, in contrast to the normal case. The final row highlights a situation, where the resulting trajectories vastly differ depending on the conditioning of the specification model.

\begin{table}[h]
\caption{The success rate of perturbing a non valid trajectory into a valid one under different user specifications.}
\begin{tabular}{cc}
\hline
User Types & Success rate \\ \hline
Careful       & 75\%     \\
Normal     & 95\%     \\
Aggressive & 100\%    \\ \hline
\end{tabular}
\label{table:backprop}
\end{table}

\subsection{Causal Analysis}

On Table.~\ref{table:causal_intervention} we can see the mean of the entailed distribution depending on the type of intervention performed. The results of Eq.~\ref{eq:neq1} can be seen in the first column under ``No intervention''. It shows the expected likelihood $E[p(v|X=x, S=s)] $ of validity of a trajectory given a set of observations with different user specifications. Conditioning on the different types of user specifications, we can see that the validity increases (from 0.43 to 1.0), meaning a higher number of possible solutions can be identified. The variety of solutions can be seen in Figure.~\ref{fig:types_traj}. This naturally follows the human assumption about the possible ways to solve a task with different degrees of carefulness. In the case of the final user type, all of the proposed trajectories have successfully solved the problem.

In the subsequent columns on Table.~\ref{table:causal_intervention} we can see the mean probability of validity for when we intervene in the world and position randomly a symbol of different type within the scene. By comparing the value with the ones in the first column (as discussed above), we can assess the inequality in Eq.~\ref{eq:neq_symbol}.

\begin{table}[h]
\caption{The respective distributions of validity p(v|X=x, S=s) with different user types depending on the intervention performed for a random trajectory to be valid under the user specification. The first column shows the mean distribution over the information obtained over the observations. The cells in bold indicate significant change with respect to the no intervention column. }
\label{table:causal_intervention}
\begin{tabular}{@{}cccccc@{}}
\toprule
User Types & No Intervention & Bowl & Plate & Cutlery & Glass \\ \midrule
Safe       &         0.43        &   \textbf{0.27}    &  \textbf{0.28}     & \textbf{ 0.31}       &    \textbf{0.30}   \\
Normal     &        0.62         &  0.62    &   0.63    &  0.62       &   \textbf{0.48}      \\
Aggressive &        1.00         &   1.00   &   1.00    &   1.00      & 1.00      \\ \bottomrule
\end{tabular}
\end{table}

In the case of a safe user specification, adding a symbol of any type decreases the probability of choosing a valid trajectory (from 0.43 down to 0.27). This indicates that the model reacts under the internalized specification to reject previously valid trajectories that interact with the intervened object.

For the normal user type, significant changes are observed only when we introduce a glass within the scene. This means it doesn't alter its behavior with respect to any of the other symbols. 

In the last case, the aggressive user type doesn't reject any of the randomly proposed trajectories and that behavior doesn't change with the intervention. It suggests the specification model, in that case, is not reacting to the scene distribution.

Based on these observations, we can postulate that the specification model has internalized rules such as ``If I want to be careful, I need to steer away from any objects on the table'' or ``To find a normal solution, look out for glass-like objects.''.

This type of causal analysis allows us to introspect in the model preference and gives us an understanding of the decision making capabilities of the model.

\section{Conclusion}

In this work, we have demonstrated how we can use a generative model to differentiate between behavioral types by using expert demonstrations. We have shown how the performance changes with the number of trajectories illustrated in a scene. Additionally, by using the same learned model, we can change any solution to satisfy the preference of a particular user type, by taking gradient steps obtained from the model.

We have shown that using causal analysis we can extract the causal link between the occurrence of specific symbols within the scene and the expected validity of a trajectory. The models exhibit different behaviors with regards to the different symbols within the scene leading us to correctly assume the underlying specifications that the humans were using during the demonstrations.

\begin{acks}
    This research is supported by the Engineering and Physical Sciences Research Council (EPSRC), as part of the CDT in Robotics and Autonomous Systems at Heriot-Watt University and The University of Edinburgh. Grant reference \grantnum{EP/L016834/1}{EP/L016834/1}.
\end{acks}


\bibliographystyle{ACM-Reference-Format}  
\balance  
\bibliography{references}  

\end{document}